\documentclass[a4paper,fleqn]{cas-sc}

\usepackage[authoryear,longnamesfirst]{natbib}
\usepackage{xcolor}
\usepackage{hyperref}
\usepackage{pdfpages}
\usepackage{placeins}
\usepackage{multirow}
\usepackage{caption}
\usepackage{subcaption}

\def\tsc#1{\csdef{#1}{\textsc{\lowercase{#1}}\xspace}}
\tsc{WGM}
\tsc{QE}
\tsc{EP}
\tsc{PMS}
\tsc{BEC}
\tsc{DE}

\begin{document}
\let\WriteBookmarks\relax
\def\floatpagepagefraction{1}
\def\textpagefraction{.001}
\shorttitle{Interpretable Classification of Circulating Tumor Cell Phenotypes}
\shortauthors{S. Su et~al.}

\title  [mode = title]{Data-Efficient and Interpretable Classification of Circulating Tumor Cell Phenotypes in Microfluidic Devices via Deep Learning}                      
\tnotemark[1]

\author[1]{Serena Su}[type=,
                        auid=000,bioid=1,
                        prefix=,
                        role=,
                       orcid=0009-0002-1026-482X]

\credit{Methodology, Software, Original draft preparation}

\affiliation[1]{organization={Department of Mathematics \& Statistic, Texas Tech University},
                city={Lubbock},
                postcode={79407}, 
                state={Texas},
                country={USA}}

\author[1]{Yifan Wang}[orcid=0000-0003-3292-4972]
\ead{yifan.wang@ttu.edu}
\cormark[1]
\credit{Conceptualization of this study, Methodology, Data curation, Original draft preparation}

\author[1]{Senwei Liang}[
   role=,
   suffix=,
   orcid=0000-0002-3558-6828
   ]
\ead{senwei.liang@ttu.edu}
\cormark[1]

\credit{Conceptualization of this study, Methodology, Software, Original draft preparation}

\cortext[cor1]{Corresponding author}

\begin{abstract}
Accurate classification of circulating tumor cell (CTC) phenotypes can provide valuable information for assessing metastatic potential. Label-free microfluidic devices provide a hydrodynamic obstacle course that transforms subtle biophysical characteristics of CTCs, including size and deformability, into distinct kinematic trajectories. However, the highly nonlinear fluid–structure interactions governing these trajectories make the inverse problem of inferring cellular phenotype from trajectory data analytically intractable. While deep neural networks (DNNs) have emerged as a powerful approach for addressing this inverse problem, their effectiveness is constrained by the limited availability of trajectory data and the lack of physical interpretability inherent to their black-box nature. 

To address these challenges, we propose an interpretable and data-efficient DNN framework for trajectory-based CTC classification. To mitigate the scarcity of data, we develop Subsequence (SubSeq), a targeted augmentation strategy that randomly extracts informative local trajectory segments during training to promote learning from localized patterns. To interpret the trained DNN, we further apply Gradient-weighted Class Activation Mapping (Selvaraju et al. 2017) to identify the trajectory features and physical regions of the microfluidic device that drive model predictions.
Experimental results demonstrate that SubSeq improves classification accuracy over the evaluated baseline and augmentation methods. Furthermore, interpretability analysis suggests that localized trajectory segments contain substantial biophysical information relevant to accurate classification. This provides justification for SubSeq and also highlights the redundancy of full-length trajectories. More broadly, the proposed framework views microfluidic geometries as physical encoders of cellular mechanical properties, providing mechanistic insights that may inform the future design of diagnostic devices.
\end{abstract}

\begin{highlights}
\item We proposed a data efficient method called SubSeq argumentation for Circulating Tumor Cell Classification.
\item We developed a Grad-CAM based analysis to identify phenotype-specific trajectory regions inside the microfluidic channel and to reveal interactions between cell, microposts and fluid that drive classification.
\item Our analysis revealed that the cell velocity captures the primary predictive signals, while position trajectory provides complementary information. Combining the two information yields the highest classification performance.
\item We found that classification relies primarily on localized trajectory segments rather than complete cell trajectories, suggesting that targeted segment analysis could reduce imaging and computational demands in future laboratory experiments.
\end{highlights}


\maketitle

\section{Introduction}
Circulating tumor cells (CTCs) are malignant cells shed from primary or metastatic tumors into the bloodstream and are widely recognized as important mediators of cancer metastasis. Detecting and classifying CTCs provide valuable insights into cancer diagnosis, prognosis, treatment response, and recurrence risk \citep{Pantel2008,AlixPanabieres2014}, making CTC analysis central to the development of liquid biopsy technologies. Beyond their mere detection in blood, comprehensive classification of CTC phenotypes is essential. CTCs exhibit substantial phenotypic heterogeneity in size, morphology, deformability, surface marker expression, and mechanical rigidity. These characteristics often reflect clinically relevant biological states, including epithelial--mesenchymal transition, metastatic potential, and therapeutic resistance~\citep{Yu2013,Joosse2015,Micalizzi2017}. 

\begin{figure}
\centering
\includegraphics[width=0.7\linewidth]{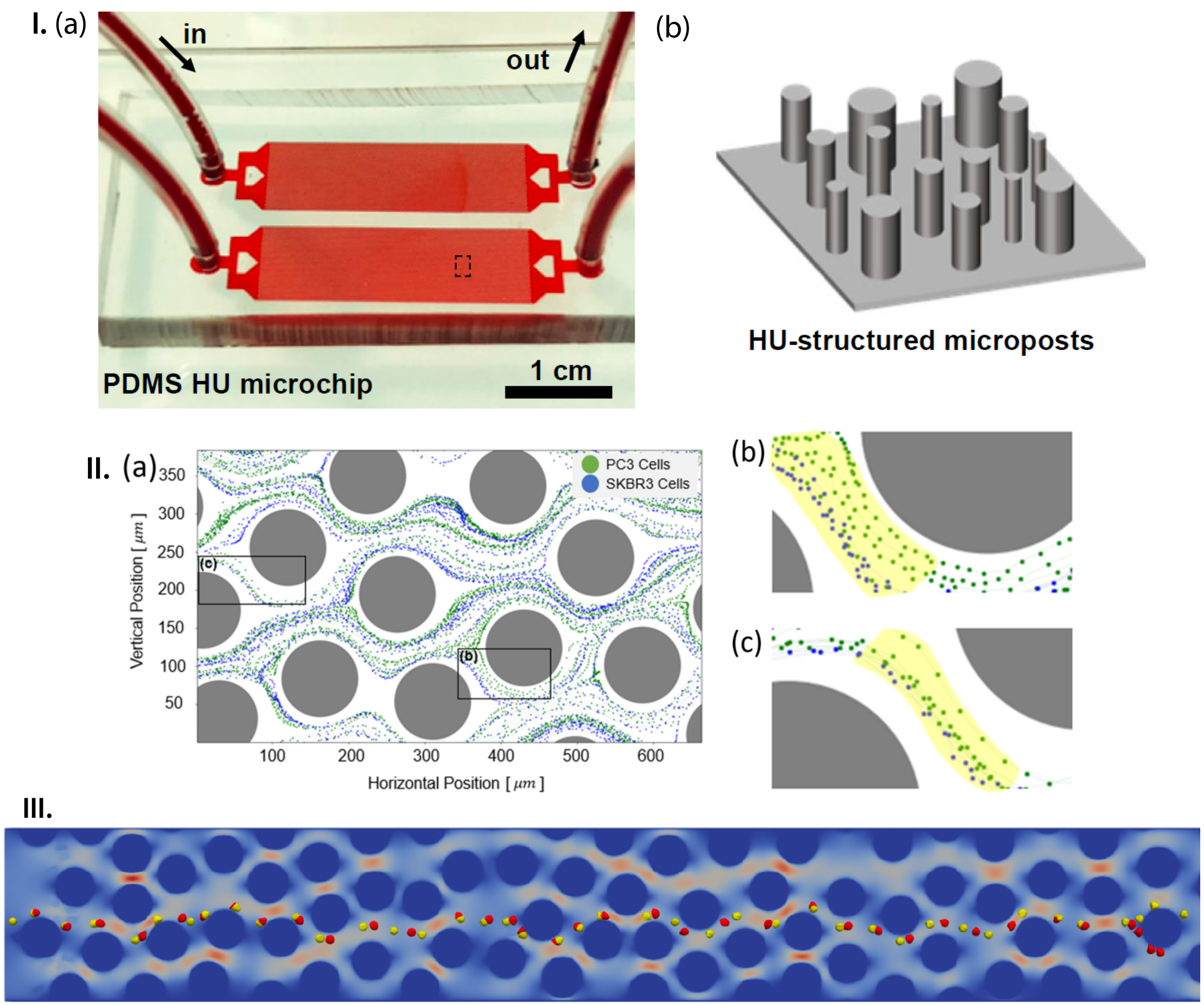}
\caption{ Subfigure I. The hyperuniform microfluidic device is shown in (a), with a schematic illustration of the micropost arrangement presented in (b). 
Subfigure II. Experimentally observed trajectories of PC3 and SKBR3 cancer cells are shown in an overall view in (a) and in two enlarged views in (b) and (c), corresponding to the regions indicated by the labeled black boxes (Images courtesy of the Wei Li laboratory). PC3 cells are shown in green, whereas SKBR3 cells are shown in blue. 
Subfigure III. The numerical simulation demonstrates distinct trajectories for two simulated circulating tumor cell types, shown in red and yellow, with different elastic properties. Both cells are released from the same initial location~\citep{Rifat2024}.}
    \label{fig:device}
\end{figure}

Traditional biochemical labeling methods can fail to classify CTC subpopulations that lack the targeted surface markers or undergo phenotypic changes during disease progression \citep{Micalizzi2017}. To overcome this limitation, label-free microfluidic devices have emerged as a powerful physics-based alternative that exploits the intrinsic biophysical properties of cells \citep{DiCarlo2009,Whitesides2006,Chen2014,Warkiani2015}. In these systems, cells are transported through microstructured channels (as shown in Figure~\ref{fig:device}a), such as arrays of stationary microposts, where they undergo continuous interactions with both the surrounding fluid and the channel structures. As a cell navigates through these microstructures, its intrinsic biophysical properties, such as size and deformability, govern its hydrodynamic response, leaving a distinctive signature on its trajectory. However, because the distributions of these properties often overlap across different cell phenotypes, the resulting trajectories exhibit highly nonlinear and coupled dynamics. Consequently, these complex motion patterns are difficult to describe using conventional analytical models or to classify with simple threshold-based sorting strategies. 

Recent studies \citep{Rifat2024,Kumar2025} have leveraged deep neural network (DNN)-based methods to validate the effectiveness of the hyperuniform microfluidic device using numerically simulated cell trajectories for CTC phenotype classification. Figure~\ref{fig:device}. I (a) and (b) show the hyperuniform microfluidic device and its micropost arrangement, which is disordered yet statistically uniform. Subfigures in Figure~\ref{fig:device}. II presents experimentally observed trajectories of PC3 and SKBR3 cancer cells, shown in green and blue, respectively. The overall and enlarged views indicate that different cell types can follow distinct paths through the micropost array. Building on these experimental observations, the aforementioned studies employed fluid--structure interaction simulations to systematically investigate phenotype-dependent interactions among cells, fluid flow, and microstructures. As shown in Figure~\ref{fig:device}. III, two CTCs with different elastic properties, despite being released from the same initial location, develop distinct trajectories through their interactions with the surrounding fluid and microposts. Given that numerical simulations provide controlled access to these coupled interactions, which can be difficult to isolate and measure experimentally at sufficient spatial and temporal resolution, the aforementioned studies used simulated trajectories to systematically characterize phenotype-dependent cell behavior. By combining these trajectories with advanced deep learning architectures, including convolutional neural networks (CNNs) and recurrent neural networks (RNNs), they demonstrated that cell trajectories contain sufficient phenotype-specific information to support CTC classification. This prior work highlights the potential of hyperuniform physical architectures to translate subtle differences in cellular mechanical properties into measurable and distinguishable trajectory signatures.

Motivated by these promising results, we seek to further improve the existing CTC classification approach by addressing two key limitations. (I): Because full fluid--structure interaction simulations are computationally expensive, the available trajectory dataset remains relatively small, which may limit model generalization. (II): These DNN models operate as black boxes, making their internal decision-making processes notoriously difficult to interpret. This lack of transparency poses a limitation in biomedical applications, where predictions must be grounded in verifiable physical mechanisms \citep{Rudin2019}. Moreover, limited interpretability potentially restricts physical insight into how trajectory patterns relate to device geometry, thereby hindering the rational design of improved microfluidic architectures.

In this study, we introduce an data-efficient and interpretable deep learning framework for CTC phenotype classification. To alleviate the limitation of scarce simulation data, we propose a novel data augmentation strategy, termed Subsequence (SubSeq) sampling, which randomly extracts continuous segments from full trajectories during training. This simple yet effective strategy significantly improves classification performance, outperforming alternative methods. Next, we apply Gradient-weighted Class Activation Mapping (Grad-CAM) to project the model's classification decisions back onto the physical trajectory space \citep{Selvaraju2017}. Interestingly, the interpretability analysis shows that the trained models do not rely on the entire trajectory for accurate prediction; instead, they concentrate on specific, highly discriminative segments within the hyperuniform device. This observation provides a physical justification for the effectiveness of SubSeq, suggesting that localized cell--flow--structure interactions contain sufficient biophysical information for phenotype discrimination, while substantial portions of the full trajectories exhibit dynamic redundancy.

The proposed framework integrates microfluidic design, trajectory generation, and interpretable learning within an encoder--decoder perspective. The hyperuniform micropost array can be viewed as a physical encoder that transforms differences in CTC mechanical properties into distinct trajectory signatures through cell interactions with the device geometry and surrounding fluid. The deep learning model then acts as a decoder that uses these trajectories to classify CTC phenotypes, while Grad-CAM maps the model's learned importance scores back onto the microfluidic geometry. This analysis improves classification performance under limited-data conditions and identifies regions of the device that may contribute to phenotype discrimination. Although further physical and experimental validation is required, these findings provide a foundation for investigating how interpretable learning could inform the future optimization of microfluidic device architectures. The main contributions of this work are summarized as follows:

\begin{itemize}
    \item We propose a trajectory-based deep learning framework with a Subsequence (SubSeq) sampling strategy and integrated Grad-CAM analysis, enabling effective learning from limited simulation data while providing interpretable mappings from trajectory segments to model predictions.
    \item We show that the cell velocity captures the primary predictive signals, while the position trajectory provides complementary information. Combining the two pieces of information yields the highest classification performance.
    \item We demonstrate that localized cell--flow--structure interactions within a hyperuniform microfluidic device contain sufficient biophysical information for CTC phenotype classification, revealing that full trajectories exhibit significant redundancy and providing mechanistic insight for the rational design of microfluidic architectures.
\end{itemize}

The remainder of this paper is organized as follows. Section~\ref{sec:relatedwork} reviews related works. Section~\ref{sec:prelim} introduces the problem setup for CTC classification. Section~\ref{sec:proposed} presents the proposed SubSeq sampling and Grad-CAM-based interpretability method. Section~\ref{sec:exp} reports experimental results, and in the end, Section~\ref{sec:conclu} concludes the paper.

\section{Related Work}
\label{sec:relatedwork}
\subsection{In Silico Microfluidic CTC Classification}

Integrating physical modeling, rational device design, and phenotypic profiling provides a clear pathway toward label-free microfluidic platforms that both detect rare cells and assess their biological significance. Exploring these frontiers, Tan et al. developed a simulation framework for CTC transport and adhesion, demonstrating how flow conditions, device geometry, ligand density, and cell properties govern capture dynamics \citep{Tan2019CTCTransportAdhesion}. To optimize this capture mechanically, Wang and Li introduced a novel device featuring hyperuniform micro-post arrangements; this structured, heterogeneous geometry creates unique cell–flow interactions that enable label-free detection and trajectory-based classification \citep{WangLi2024MicrofluidicDeviceCTCs}. Expanding on these analytical capabilities, Joshi et al. reviewed how microfluidic platforms are moving beyond simple isolation to characterize clinically relevant phenotypes, such as cell size, deformability, adhesion, metabolism, and surface markers \citep{Joshi2025MicrofluidicTumorCellPhenotyping}.

\subsection{Data Augmentation for DNN}
Data augmentation improves machine learning model generalization by artificially expanding training set diversity. It is important in data-scarce environments, where it mitigates overfitting by exposing models to a broader spectrum of input variations. In computer vision, standard image augmentations include geometric transformations (e.g., rotation, flipping, translation~\citep{he2016deep}) and pixel-level perturbations like noise injection. Recently, advanced regularization techniques such as Cutout~\citep{cutout} and Mixup~\citep{mixup,lin2024continuous}, alongside automated search strategies like AutoAugment~\citep{Cubuk_2019_CVPR} and data engine~\citep{liang2024aide}, have further advanced visual representation learning.

Migrating image augmentations to time-series data requires accounting for strict temporal dependencies, as arbitrary transformations can easily distort sequential dynamics and generate invalid samples. Consequently, specialized techniques are categorized into time-domain, frequency-domain~\citep{gao2020robusttad}, and advanced learning-based approaches~\citep{wen2021time}. Within the time domain, methods manipulate raw sequences directly: window cropping extracts local segments to multiply samples~\citep{CHEN2021126}, flipping reverses numerical signs for symmetric datasets, window warping alters local dynamics via temporal compression or extension~\citep{wen2021time}, and noise injection adds minor variations without altering underlying semantic labels~\citep{wen2019time}.

\subsection{DNN Interpretability}
Understanding the decision-making process of DNNs is an important aspect of model evaluation, especially in domains such as medical diagnosis~\citep{huang2025generic}. In computer vision, Class Activation Maps (CAMs) are widely used to explain model predictions by highlighting image regions that contribute most to a target class. The original CAM method proposed by \citep{zhou2016} generates class-specific activation maps by linearly combining the final convolutional feature maps with class-specific weights. Since then, numerous CAM variants have been developed, broadly categorized into gradient-free methods (e.g., CAM, Score-CAM, Recipro-CAM) and gradient-based methods (e.g., Layer-CAM, Grad-CAM) \citep{minh2023}. Among these, Gradient-weighted CAM (Grad-CAM) \citep{gradcam} is one of the most widely adopted approaches. It has also been extended to time-series analysis, where it identifies important temporal segments, with successful applications in national demand forecasting \citep{VANZYL2024122079} and manufacturing anomaly detection \citep{hyun2024}.

\section{Preliminary}
\label{sec:prelim}
This section introduces the trajectory-based classification task for circulating tumor cell phenotypes and the convolutional neural network (CNN) used in the proposed framework.
\subsection{CTC Trajectory and Phenotypes}

Cell trajectories in the hyperuniform microfluidic device were generated using numerical simulations of a coupled fluid--structure interaction model~\citep{Rifat2024}. The surrounding blood plasma was modeled as an incompressible Newtonian fluid governed by the Navier--Stokes equations and solved using the lattice Boltzmann method. Each CTC membrane was represented by a spring-network model whose deformation was computed using the finite element method. The spring-network accounts for stretching, bending, and surface area and volume conservation effects, with its dynamics coupled to the fluid through immersed boundary method.

The dataset consists of $N$ trajectory samples $\mathcal{D}=\{(\mathbf{X}^{(i)}, p^{(i)})\}_{i=1}^{N}$,
where
$
\mathbf{X}^{(i)}=\left\{(x_t^{(i)},y_t^{(i)},z_t^{(i)},v_{x,t}^{(i)},v_{y,t}^{(i)},v_{z,t}^{(i)})\right\}_{t=1}^{T}
$
represents the position (first three coordinates) and velocity (last three coordinates) time series of the $i$-th cell, and $p^{(i)}\in\{0,1\}$ denotes its phenotype label, corresponding to soft and hard CTC types, respectively. The trajectory-based CTC phenotype classification task aims to learn a classifier that maps $\mathbf{X}^{(i)}$ to its corresponding CTC phenotype.

However, generating cell trajectories is computationally demanding because each trajectory requires a complete time-dependent fluid--structure interaction simulation. As a result, only a limited number of training samples are available, motivating the use of effective data augmentation strategies to improve the generalization of DNN models.

\subsection{CNN for CTC Classification}

\begin{figure}
    \centering
    \includegraphics[width=0.9\linewidth]{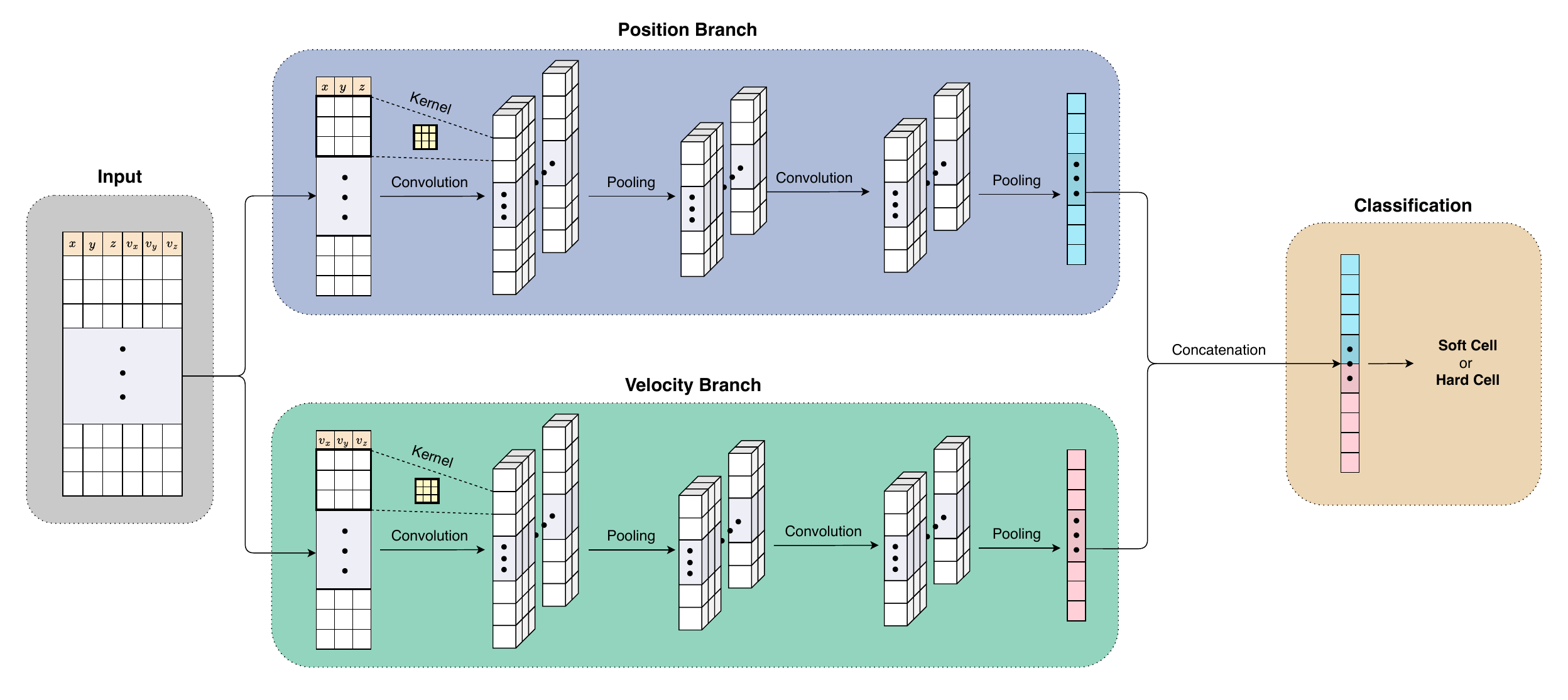}
    \caption{Convolutional neural network for CTC classification~\citep{Kumar2025}. The model processes the position and velocity inputs through parallel branches, concatenates the extracted features, and finally performs classification using a multilayer perceptron to produce the final phenotype classification.}
    \label{fig:architecture}
\end{figure}

CNNs have demonstrated promising performance in trajectory-based CTC phenotype classification~\citep{Kumar2025}. In this study, we adopt the CNN architecture illustrated in Figure~\ref{fig:architecture} as the primary testbed for evaluating the proposed framework. Specifically, we use this architecture to quantify the effect of data augmentation on classification performance relative to the baseline training procedure and apply Grad-CAM to interpret the features that influence the trained model's predictions. Although the present analysis focuses on this CNN architecture, the proposed framework can also be applied to other neural network models for sequential data.

As shown in Figure~\ref{fig:architecture},  the input trajectory is split into two sequences: the positional sequence $(x_t,y_t,z_t)_{t=1}^T$ and the velocity sequence $(v_{x,t},v_{y,t},v_{z,t})_{t=1}^T$. These two sequences are processed by two parallel branches with the same architecture. Each branch takes a three-channel time series as input and applies a one-dimensional convolutional layer with 256 filters of kernel size 3, followed by a ReLU activation and a max-pooling layer with kernel size 2. A second one-dimensional convolutional layer with 256 filters and kernel size 3 is then applied, again followed by a ReLU activation. Adaptive average pooling subsequently aggregates the temporal dimension into a fixed-length 64-dimensional feature vector. Finally, the feature vectors from the position and velocity branches are concatenated and fed into a multilayer perceptron to produce the final phenotype classification. 

\section{Proposed Methods for Improving and Interpreting CTC Classification}
\label{sec:proposed}
This section introduces the proposed \textit{SubSeq} sampling method, a data augmentation technique designed to improve CTC classification performance, and \textit{Grad-CAM} analysis, which is used to interpret the trained models and identify salient features of CTC trajectories.

\subsection{Subsequence (SubSeq) Augmentation}
Inspired by the intuition behind Cutout~\citep{cutout}, SubSeq intentionally limits the trajectory information available to the classifier during training. By exposing only a randomly selected temporal segment, the model is encouraged to learn discriminative local motion patterns rather than relying on complete trajectories. 

Given a trajectory sample
$\mathbf{X}^{(i)}
=
\left\{
(x_t^{(i)},y_t^{(i)},z_t^{(i)},
v_{x,t}^{(i)},v_{y,t}^{(i)},v_{z,t}^{(i)})
\right\}_{t=1}^{T}$, the proposed SubSeq method generates an augmented training input sample by randomly extracting a contiguous temporal segment from the original trajectory. Let \(r \in (0,1]\) denote the minimum subsequence ratio. The minimum subsequence ratio \(r\) controls the amount of information retained in each sample. The minimum allowable subsequence length is defined as $
L_{\min}
=
\max\!\left(
\left\lfloor rT \right\rfloor,
1
\right)
$, where $\lfloor\cdot\rfloor$ denotes the floor operator, i.e., the greatest integer less than or equal to its argument. 
A subsequence length \(L\) is then randomly and uniformly sampled from $
L
\sim
\mathcal{U}
\left(\{L_{\min},L_{\min}+1,\cdots,T\}\right),$ and a starting index \(s\) is uniformly selected from $s
\sim
\mathcal{U}
\left(
\{1,2,\ldots,T-L+1\}
\right).
$

The extracted subsequence is $
\mathbf{X}^{(i)}_{\mathrm{sub}}
=
\left\{
\mathbf{u}^{(i)}_t
\right\}_{t=s}^{s+L-1},$ where $
\mathbf{u}^{(i)}_t
=
(x_t^{(i)},y_t^{(i)},z_t^{(i)},
v_{x,t}^{(i)},v_{y,t}^{(i)},v_{z,t}^{(i)}).$ To maintain a fixed input length for the CNN, the extracted subsequence is padded with zeros to recover the original trajectory length \(T\), e.g., 
$
\widetilde{\mathbf X}^{(i)}
=
\left(
\mathbf{u}^{(i)}_s,
\mathbf{u}^{(i)}_{s+1},
\ldots,
\mathbf{u}^{(i)}_{s+L-1},
\underbrace{\mathbf 0,\ldots,\mathbf 0}_{T-L}
\right).
$ Consequently, the classifier is trained using only a randomly selected portion of each trajectory while preserving a consistent input size.

\subsection{Importance Score Visualization via Grad-CAM}

For interpretability analysis, we employ Grad-CAM to identify the trajectory features that are most influential to the trained model predictions. Grad-CAM leverages the gradients obtained through backpropagation of the target output with respect to intermediate feature maps to generate a coarse localization map to highlight the regions that contribute most to the prediction. Since the CNN as shown in Figure~\ref{fig:architecture} contains separate branches for processing position and velocity inputs, Grad-CAM is applied to the final convolutional layer of either the position or velocity branch.

Given a target class $c$, Grad-CAM generates a class-discriminative localization map $\mathrm{ReLU}
\left(
\sum_k a_k^c A^k
\right)$, where $A^k$ denotes the $k$-th feature map of the last convolutional layer and $a_k^c$ represents its importance weight for class $c$. The weights are computed by globally averaging the gradients of the class logit $y^c$ with respect to the feature map activations, $a_k^c = \frac{1}{Z}
\sum_{i,j}
\frac{\partial y^c}{\partial A_{ij}^{k}}$, where $Z$ is the total number of elements in the feature map. The resulting weights quantify the contribution of each feature map to the prediction of class $c$. A weighted combination of the feature maps is then computed to produce a coarse localization map. Finally, the ReLU operation retains only positive contributions, highlighting the regions that positively influence the prediction while suppressing those that negatively affect the target class. The resulting Grad-CAM map, $L_{\mathrm{Grad\text{-}CAM}}^{c}\in\mathbb{R}^{T}$, identifies the temporal regions that are most influential to the classifier decision. Applying Grad-CAM to each branch separately produces a one-dimensional importance score for every time step in that branch. Consequently, the position and velocity branches each generate a temporal importance map, totally two time series of feature importance values.

To project the temporal importance scores into physical space, we associate the importance score at each time step with the corresponding position or velocity. Specifically, we define a bounding box enclosing the trajectory domain and partition it into a uniform 2D spatial grid. The importance scores from all test trajectories are then averaged within the corresponding grid boxes according to their spatial coordinates. The resulting spatial importance heatmap captures the statistical distribution of phenotype-discriminative regions across the physical domain.

\section{Experiments}
\label{sec:exp}

In this section, we first validate the performance of SubSeq augmentation and then leverage Grad-CAM for model interpretation. For performance evaluation, we benchmark SubSeq across diverse data regimes (from data-rich to data-scarce) and compare it with other data augmentation methods. For interpretability analysis, we first examine the individual contributions of position versus velocity features, and then statistically aggregate the test-set Grad-CAM maps to identify the generalized spatial and kinematic regions that dictate the classification decisions.

\subsection{Experimental Protocol}
\label{sec:protocol}
The dataset contains a total of 516 CTC trajectories, including 262 soft-cell and 254 hard-cell trajectories. Among them, 416 trajectories are used for training and 100 for testing. For the Grad-CAM analysis, we further consider a filtered subset of 50 test trajectories that traverse the entire microfluidic channel, consisting of 36 soft-cell and 14 hard-cell trajectories. 

The CNNs were trained using the cross-entropy loss function and the Adam optimizer~\citep{kingma2014adam}. To ensure the statistical reliability of the results against variations in data partitioning and NN model initialization, we evaluated each configuration using 75 independent runs. For each fixed train-test ratio, these runs were distributed across 5 distinct train-test splits, with each split evaluated across 15 independent trials. Each run was trained for 2000 epochs, and the final results from all 75 runs were aggregated and visualized using boxplots to evaluate the model performance.

Model performance is evaluated using testing accuracy and the area under the receiver operating characteristic curve (ROC-AUC). The ROC-AUC is calculated on a held-out test set to assess the model's overall discriminative ability, providing a threshold-independent measure of classification performance. The ROC-AUC metric ranges from 0.5 to 1, where values closer to 1 indicate better classification performance.

\subsection{Classification Performance with SubSeq}
\label{sec:subseq_perfomance}
\paragraph{SubSeq across diverse data regimes.}
We first evaluate the impact of the SubSeq  across diverse data regimes, ranging from data-rich to data-scarce scenarios, by comparing accuracy and ROC-AUC over varying train-test splits. 

\begin{figure}
    \centering
    \includegraphics[width=0.49\linewidth]{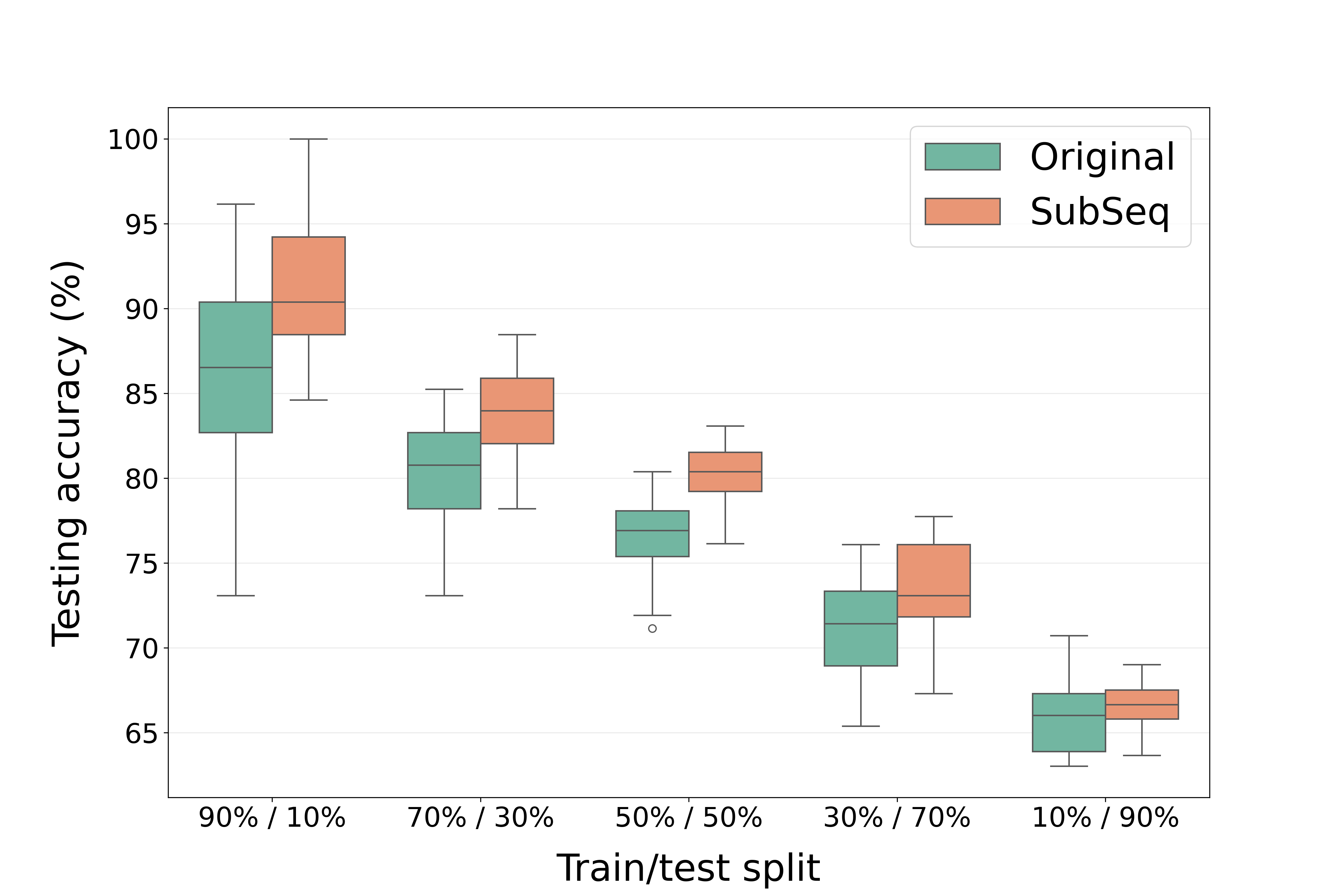}
    \includegraphics[width=0.49\linewidth]{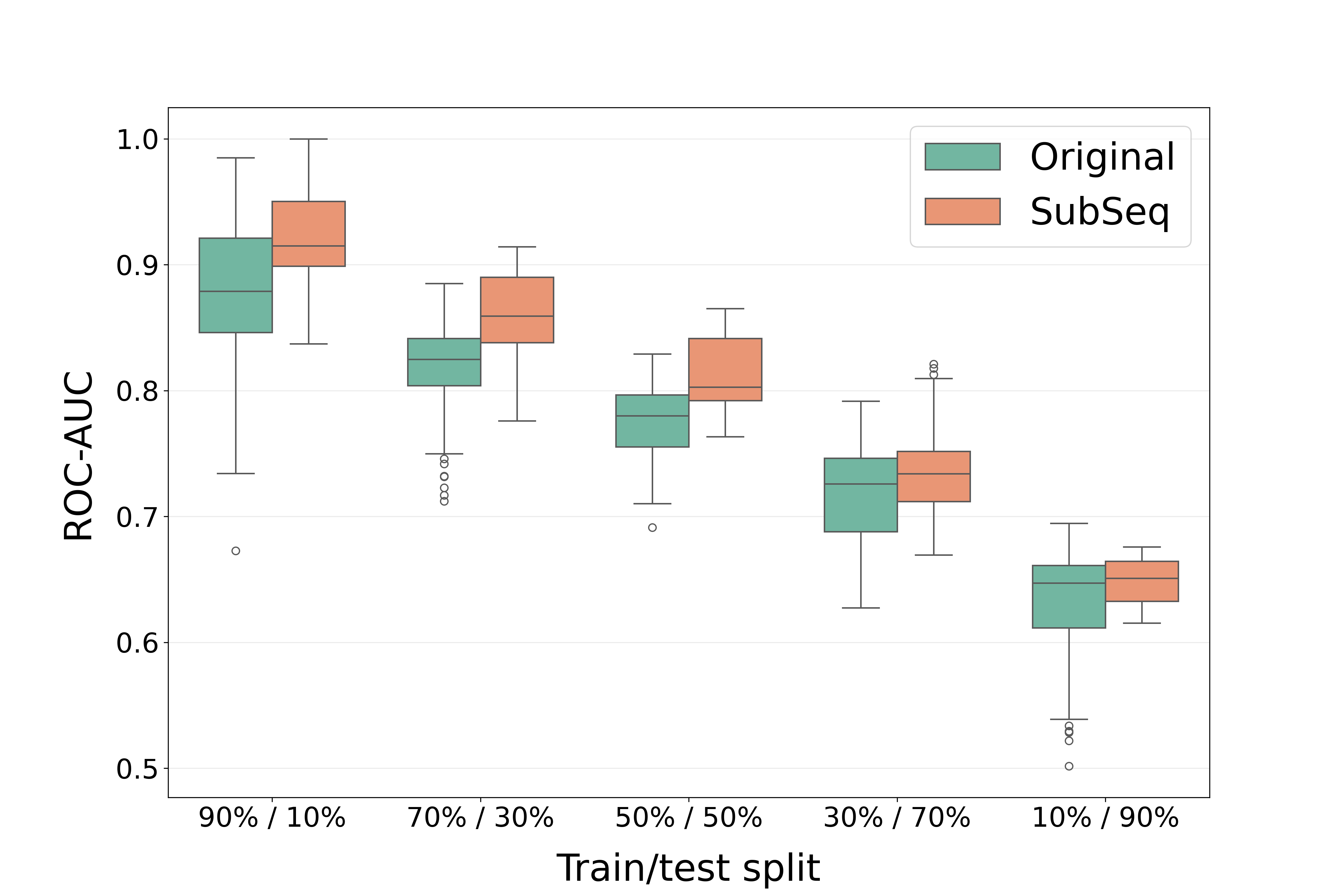}
    \caption{Comparison of original and SubSeq model performance across varying train-test splits. Boxplots show the distribution of testing accuracy (left) and ROC-AUC (right), with paired boxes indicating models trained on identical data splits. The central line denotes the median, boxes indicate the interquartile range, whiskers represent the data range, and hollow circles indicate outliers. Results show that SubSeq consistently achieves higher median testing accuracy and ROC-AUC than the baseline.
    }
    \label{fig:subseq_vs_original}
\end{figure}

Figure \ref{fig:subseq_vs_original} shows that while reducing training data expectedly degrades the performance of both models, the SubSeq model consistently outperforms the baseline. Across diverse data regimes, SubSeq delivers higher median ROC-AUC and testing accuracy. Additionally, the narrower spread of performance distributions of SubSeq reflects greater stability across independent trials. The performance gap in favor of SubSeq is particularly notable at lower training data fractions, which confirms that SubSeq mitigates overfitting and enhances generalization in data-scarce scenarios. 

\paragraph{SubSeq against alternative data augmentation methods.}

We evaluate the effectiveness and stability of SubSeq by comparing it to alternative data augmentation methods, including Cutout~\citep{cutout} and Mixup~\citep{mixup}. We provide implementation details in Appendix~\ref{appendix:imp}.

Figure \ref{fig:subseq_cutout_mixup} compares the testing accuracy and ROC-AUC across the three augmentation methods. For testing accuracy (left), SubSeq achieves a comparable median to Cutout but exhibits a significantly tighter distribution and a much higher lower bound, which mitigates the worst-case performance seen in Cutout. In terms of ROC-AUC (right), while Cutout yields the highest peak and median values, it also shows the largest variability. In contrast, SubSeq maintains a competitive median with reduced spread. Overall, while Cutout offers strong peak potential, SubSeq provides the most consistent and stable performance improvements across both metrics, making it a more reliable data augmentation strategy for the CTC dataset.

\begin{figure}
    \centering
    \includegraphics[width=0.32\linewidth]{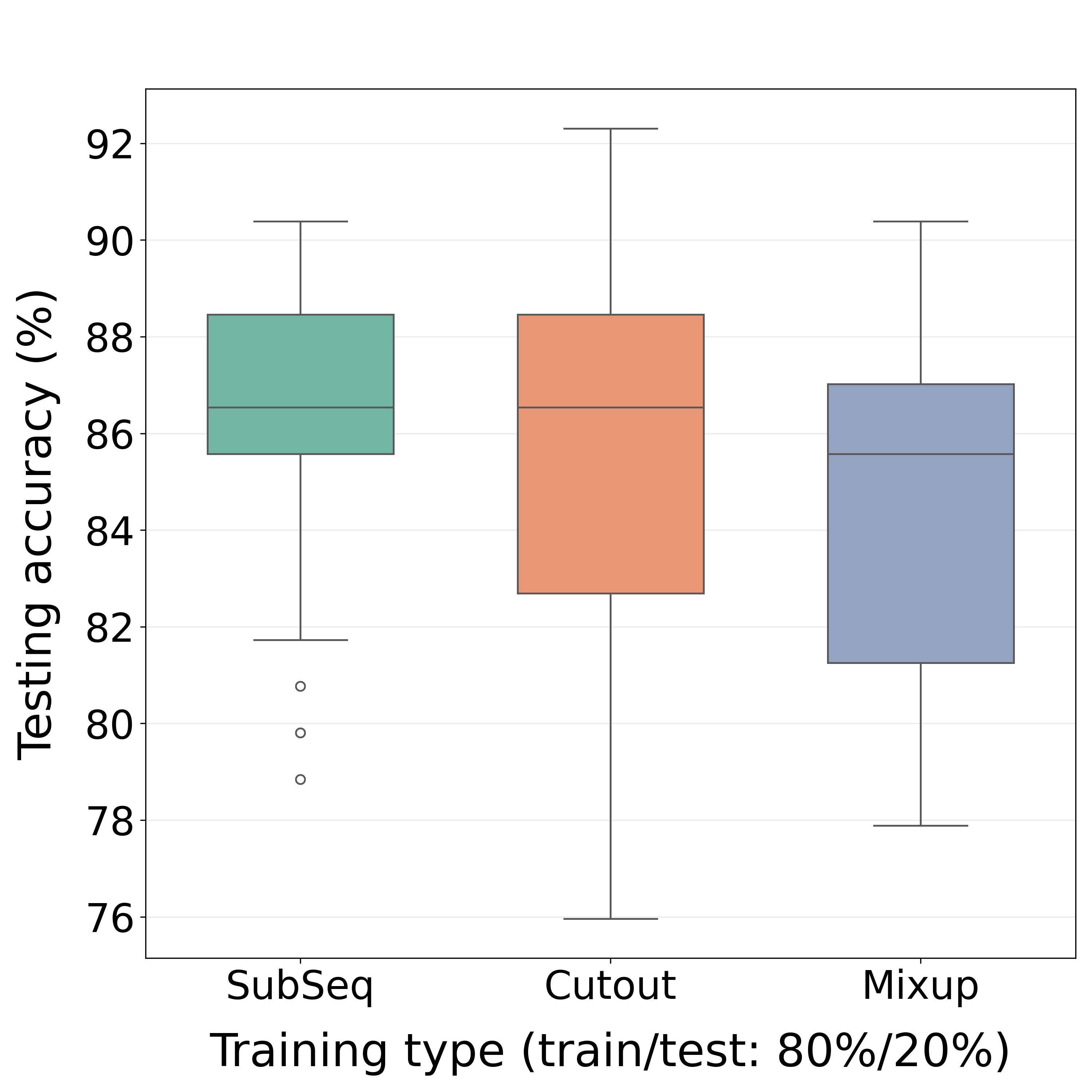}
    \includegraphics[width=0.32\linewidth]{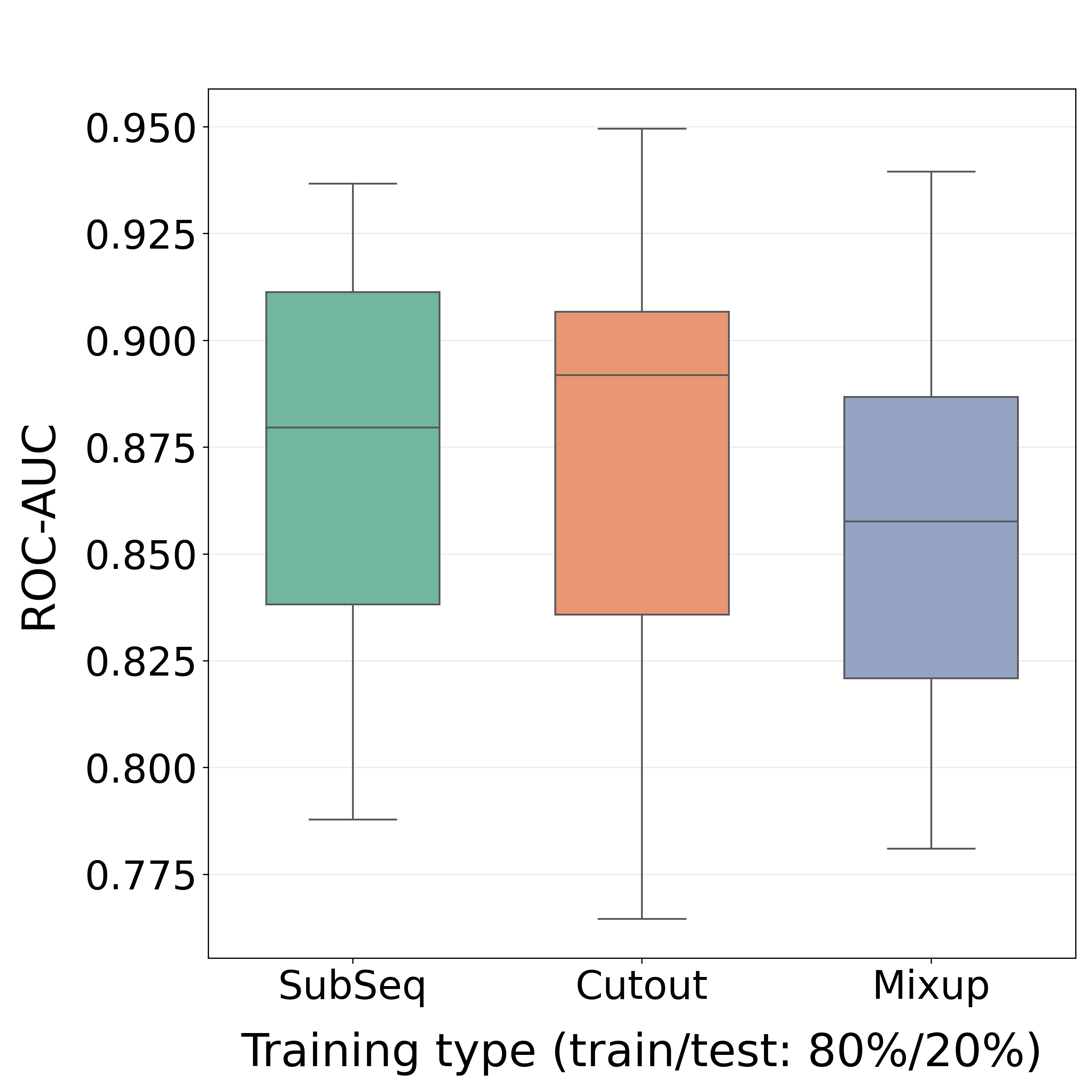}
    
    \caption{Performance comparison of data augmentation strategies under an 80\%/20\% training/testing split. Boxplots display the distribution of testing accuracy (left) and ROC-AUC (right) across the SubSeq, Cutout, and Mixup methods.
    Results show that SubSeq offers the most stable performance (lower variability) in both testing accuracy and ROC-AUC across the evaluated augmentation methods.}
    \label{fig:subseq_cutout_mixup}
\end{figure}

\subsection{Interpreting CTC Classification}
\paragraph{Effect of input features: position vs. velocity.} To gain a better understanding of the role of different input features in model performance, we evaluate the respective contributions of positional and velocity information. Specifically, we conduct an ablation study comparing a baseline configuration that utilizes both position and velocity features against a modified variant that relies solely on velocity or position data. These models are trained using the same parameters outlined in section \ref{sec:protocol} and tested across varying train-test splits. 

\begin{figure}
    \centering
    \includegraphics[width=0.45\linewidth]{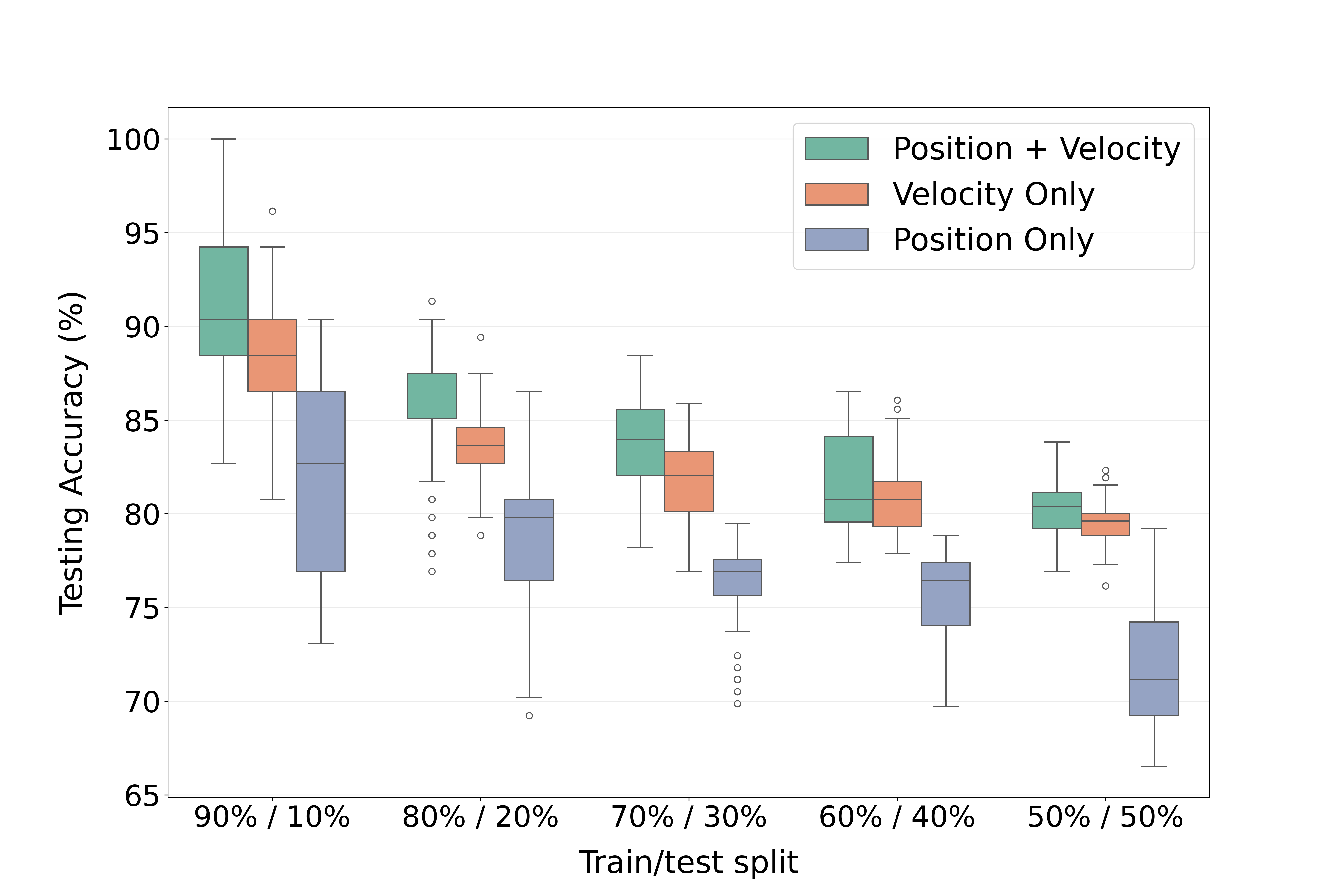}
    \includegraphics[width=0.45\linewidth]{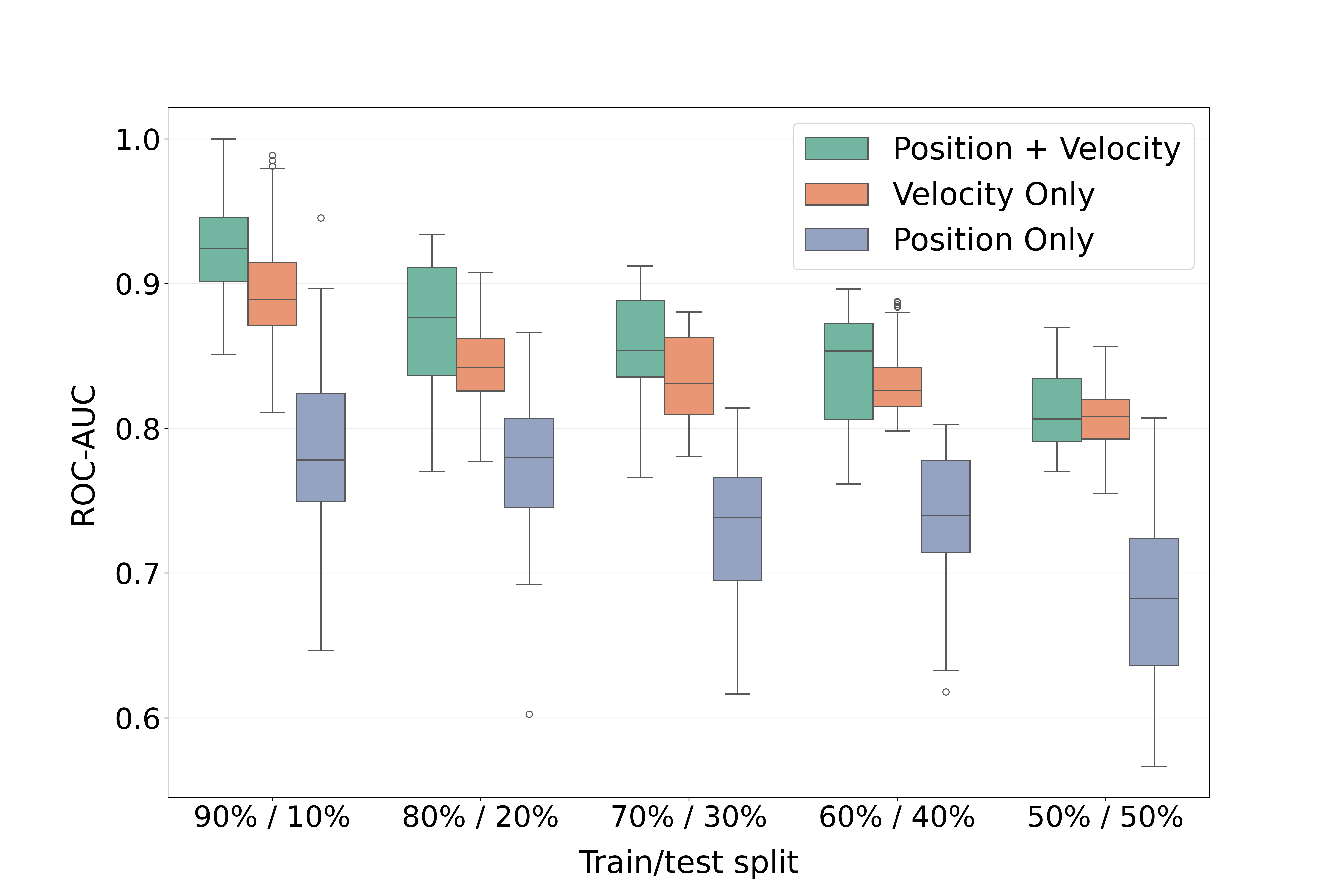}
    \caption{Effect of input representation on model performance. Boxplots display the distributions of testing accuracy (left) and ROC-AUC (right) for models trained using SubSeq with velocity-only/position-only inputs or combined position and velocity inputs across varying train--test splits.
    Results show that velocity provides most of the discriminative information, while combining position and velocity yields the best performance.}
    \label{fig:subseq_velocity_only}
\end{figure}

Figure \ref{fig:subseq_velocity_only} evaluates model performance using velocity-only or positional-only inputs versus combined position and velocity features. The velocity-only configuration delivers highly competitive results, indicating that velocity features drive the core predictive performance. Conversely, models relying solely on positional information perform significantly worse, proving that position alone is insufficient for accurate classification. Nevertheless, integrating both modalities consistently yields the best overall results across all metrics. This demonstrates that positional data provides  complementary information that enhances the model predictive capability.

\begin{figure}
    \centering
    \begin{subfigure}{0.4\linewidth}
        \includegraphics[width=\linewidth]{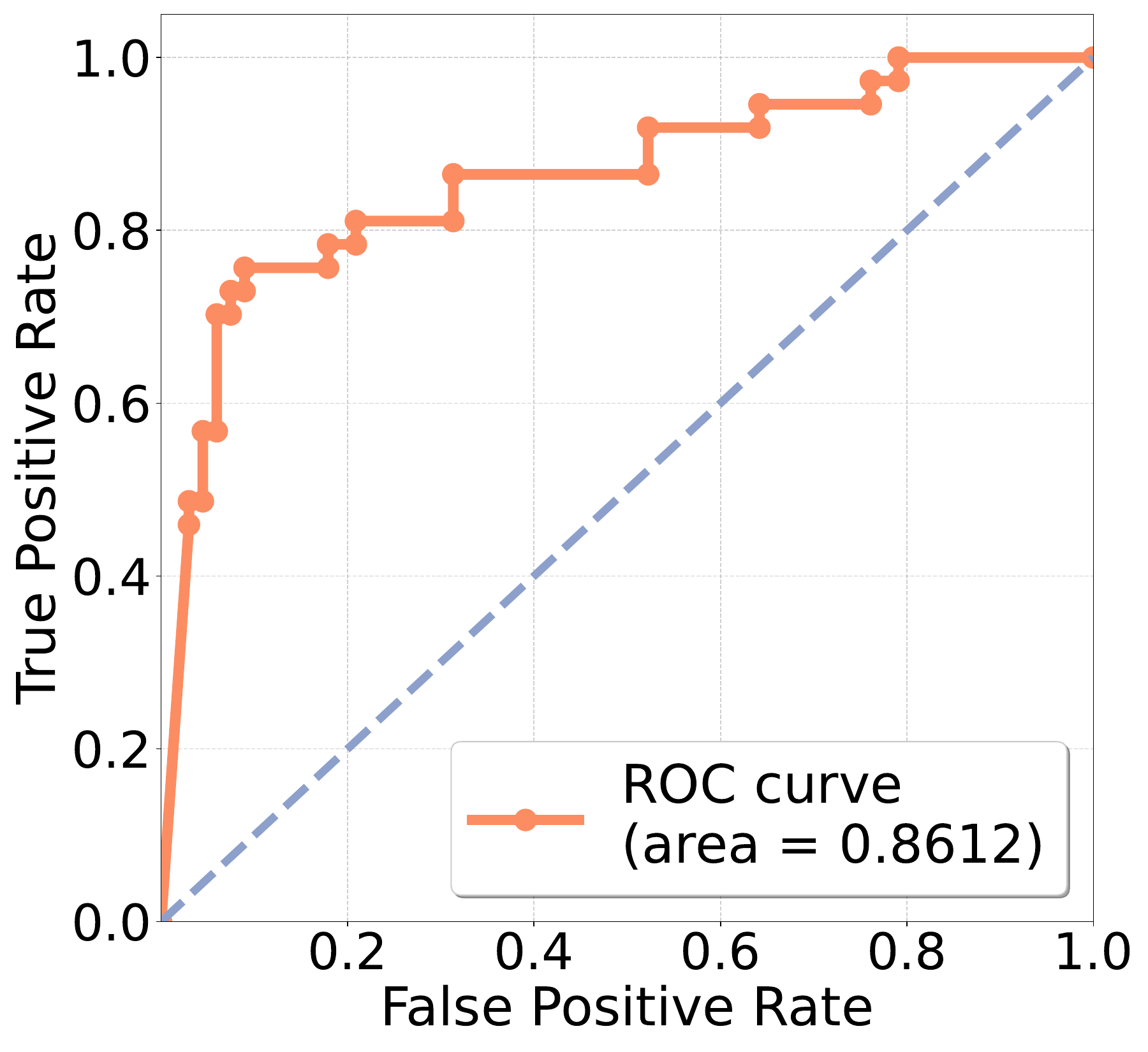}
        \caption{Position + Velocity}
    \end{subfigure}
    \begin{subfigure}{0.4\linewidth}
        \includegraphics[width=\linewidth]{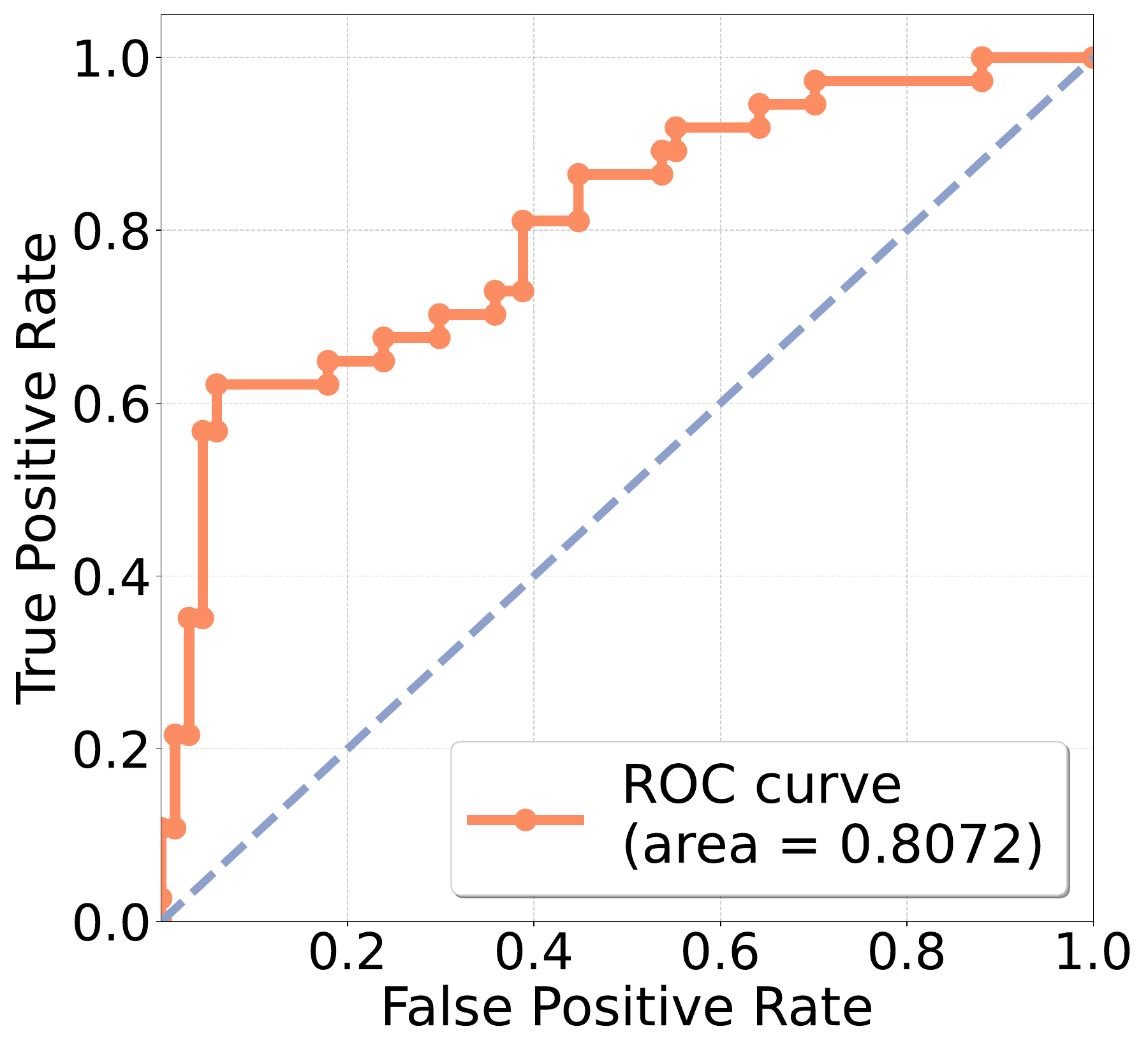}
        \caption{SubSeq \&  Position only}
    \end{subfigure}
    \begin{subfigure}{0.4\linewidth}
        \includegraphics[width=\linewidth]{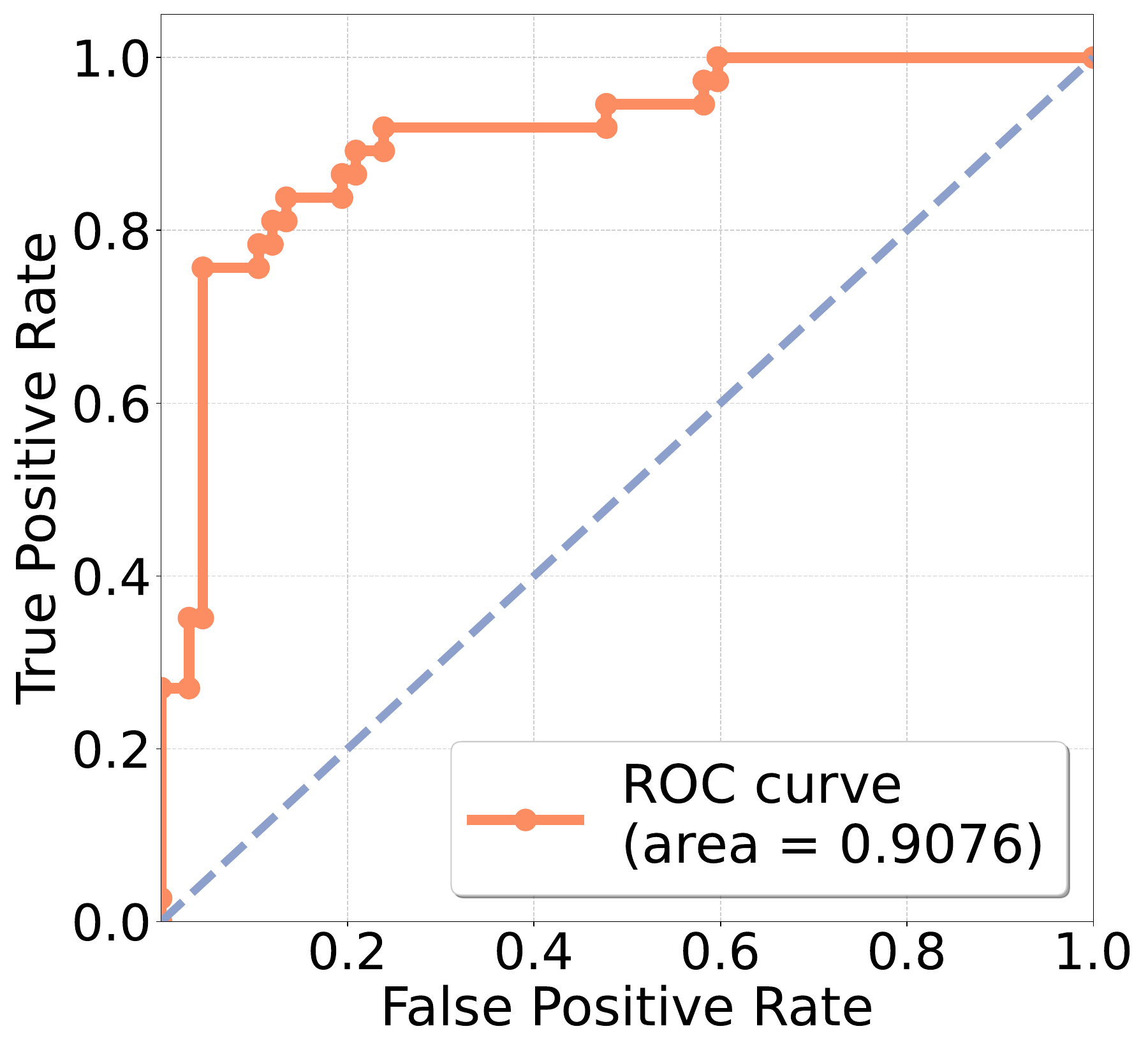}
        \caption{SubSeq \& Velocity only}
    \end{subfigure}
    \begin{subfigure}{0.4\linewidth}
        \includegraphics[width=\linewidth]{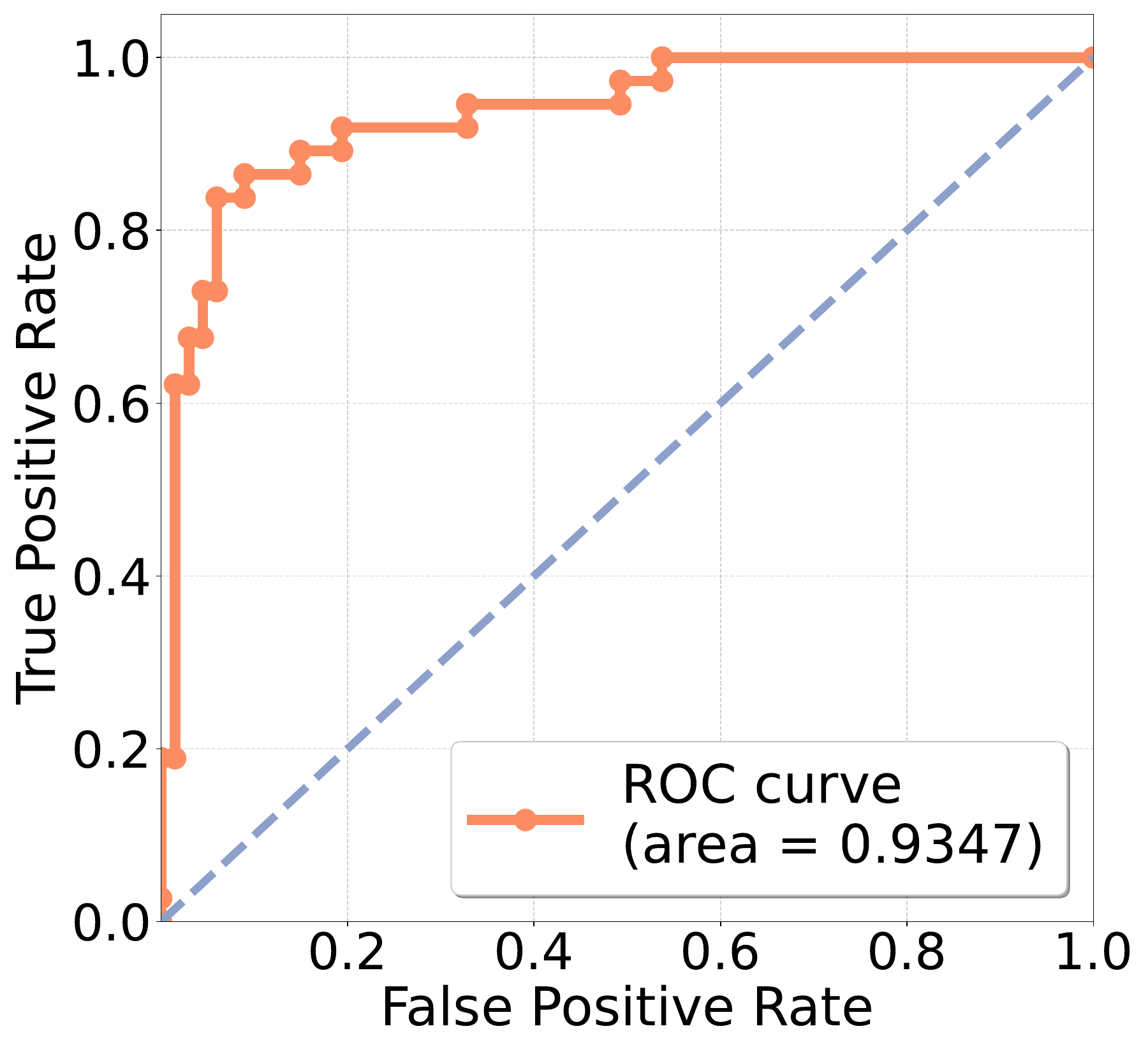}
        \caption{SubSeq \& Position + Velocity}
    \end{subfigure}
    \caption{ROC-AUC under an 80/20 train–test split across different input and augmentation configurations. Panels illustrate models trained with: (Top Left) combined position and velocity inputs without data augmentation, (Top Right) Position-only inputs with SubSeq, (Bottom Left) Velocity-only inputs with SubSeq, and (Bottom Right) combined position and velocity inputs with SubSeq. Results show that SubSeq improves the ROC-AUC from 0.8612 to 0.9347 when both position and velocity are used. We can see that velocity provides more discriminative information than position, while combining both features achieves the best classification performance.}
    \label{fig:roc_graphs}
\end{figure}

This behavior is further illustrated by the ROC curves shown in Figure \ref{fig:roc_graphs}. Quantitatively, the unaugmented baseline model utilizing combined position and velocity inputs yields an ROC-AUC of 0.8612. Introducing data augmentation boosts the performance to an AUC of 0.9076 for the velocity-only configuration and to a peak of 0.9347 for the full model incorporating both feature modalities. Visually, the velocity-only model exhibits a more rigid, stepwise curve profile compared to the smoother curve profile that includes both position and velocity inputs. This suggests that relying only on velocity data while omitting positional information leads to more discrete decision boundaries.

These findings indicate that velocity captures the essential characteristics needed for classification, whereas positional information acts as an important supplement. Combining both feature types is therefore vital to maximizing overall predictive  capability.

\paragraph{Importance score interpretation.}

\begin{figure}
    \centering
    \begin{tabular}{m{0.63\linewidth} m{0.05\linewidth} p{0.25\linewidth}}
     \raisebox{-8mm}{\includegraphics[width=\linewidth]{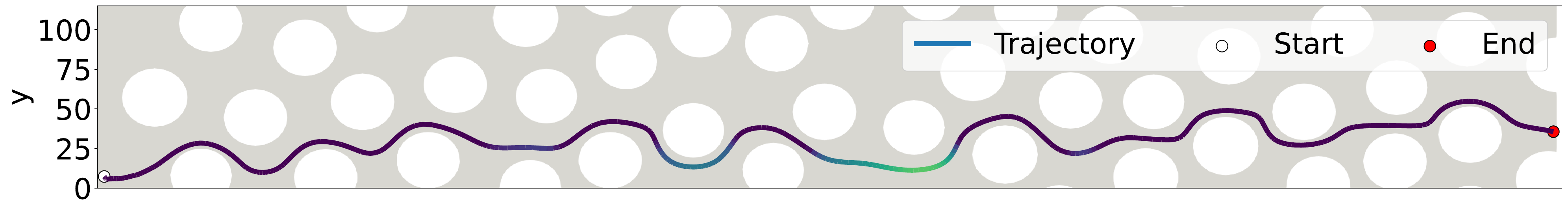}} & \raisebox{2mm}{\multirow{4}{*}{\includegraphics[width=\linewidth]{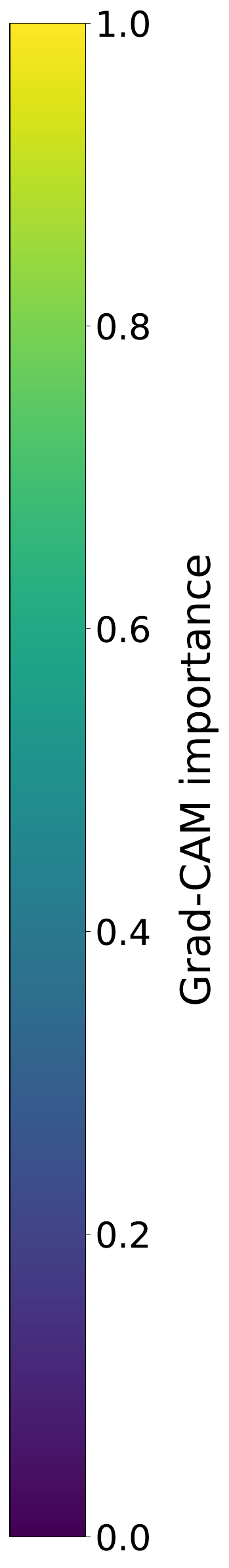}}} & \textbf{(a)} Soft Cell, Position Branch, Correct, Probability:1 \\
     \raisebox{-8mm}{\includegraphics[width=\linewidth]{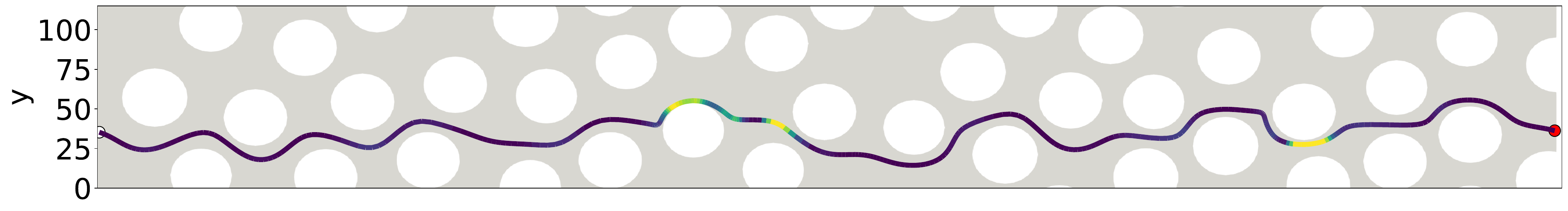}}& & \textbf{(b)} Soft Cell, Velocity Branch, Correct, Probability: 1 \\
     \raisebox{-8mm}{\includegraphics[width=\linewidth]{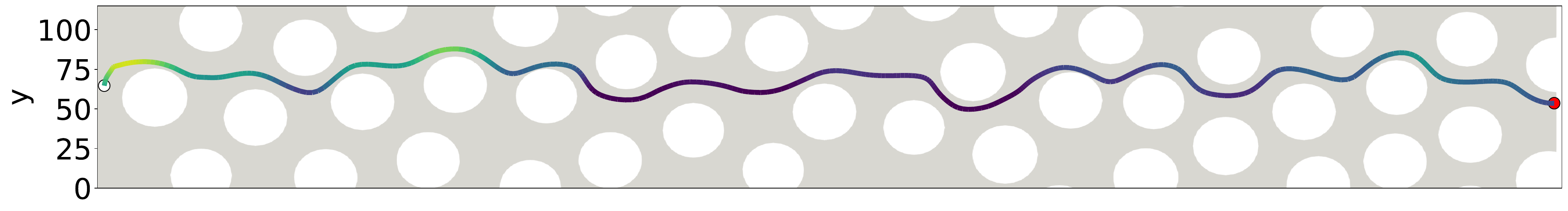}}& & \textbf{(c)} Hard Cell, Position Branch, Correct, Probability: 0.975 \\
     \raisebox{-8mm}{\includegraphics[width=\linewidth]{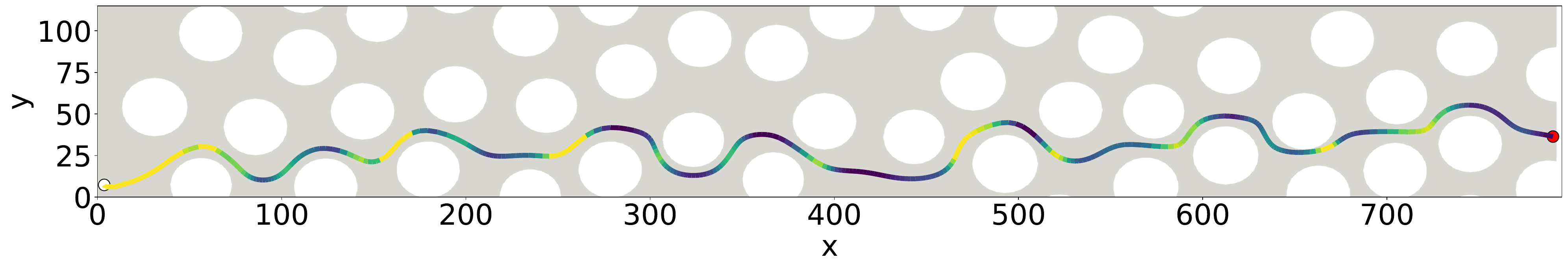}}& & \textbf{(d)} Hard Cell, Velocity Branch, Incorrect, Probability: 0.873
    \end{tabular}
    \caption{Grad-CAM importance maps for several representative test trajectories under the soft- and hard-cell prediction targets, for each input modality (position or velocity). Lighter colors indicate larger Grad-CAM importance scores. The labels on the right denote the specific conditions and outcomes for each trajectory. For example, label (a) indicates a Soft Cell trajectory analyzed via the Position Branch, which the model correctly classified with a probability of 1.0.}
    \label{fig:trajectories}
\end{figure}

\begin{figure}
    \centering
    \begin{tabular}{m{0.7\linewidth} m{0.055\linewidth} p{0.13\linewidth}}
     \raisebox{-8mm}{\includegraphics[width=\linewidth]{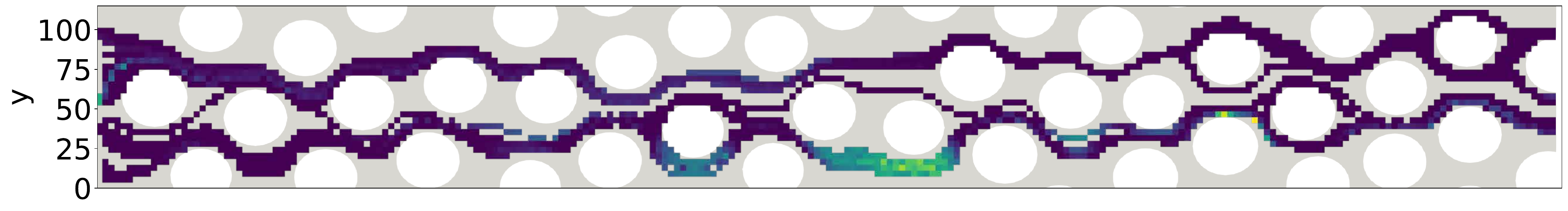}} & \raisebox{3mm}{\multirow{4}{*}{\includegraphics[width=\linewidth]{images/colorbar.pdf}}} & \textbf{(a)} Soft Cells, Position Branch \\
     \raisebox{-8mm}{\includegraphics[width=\linewidth]{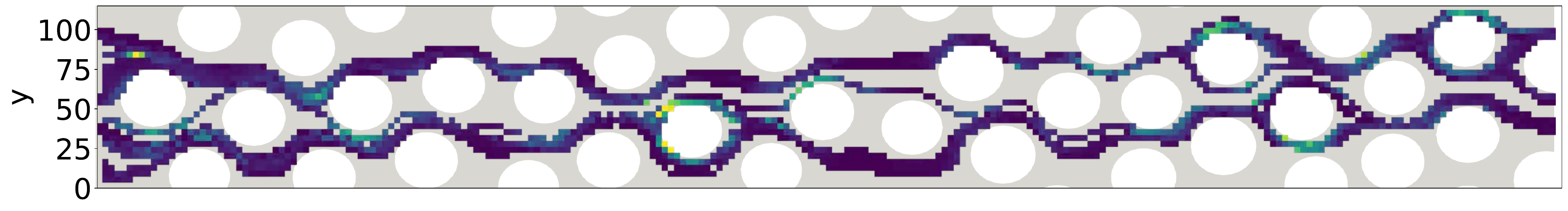}}& & \textbf{(b)} Soft Cells, Velocity Branch \\
     \raisebox{-8mm}{\includegraphics[width=\linewidth]{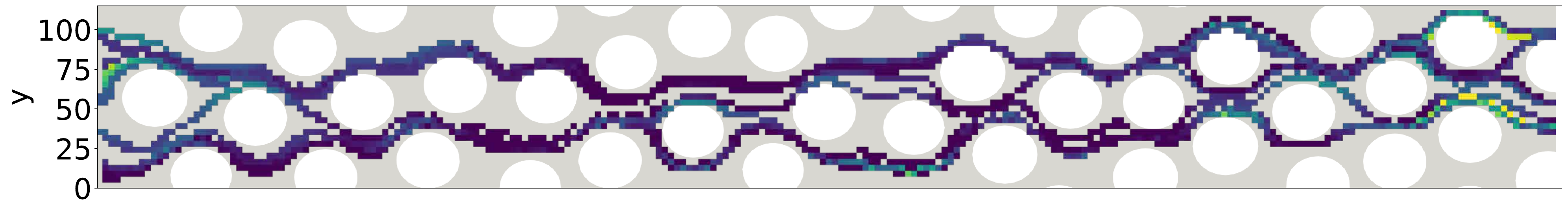}}& & \textbf{(c)} Hard cells, Position Branch \\
     \raisebox{-8mm}{\includegraphics[width=\linewidth]{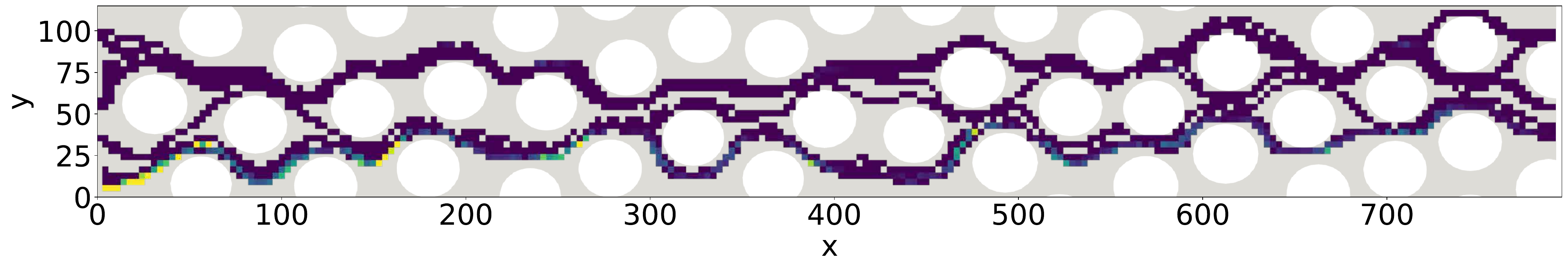}}& & \textbf{(d)} Hard cells, Velocity Branch
    \end{tabular}
    \caption{Grad-CAM importance maps aggregated over test trajectories under the soft- and hard-cell prediction targets, for each input modality (position and velocity). Lighter colors indicate larger Grad-CAM importance scores. The labels on the right indicate the conditions for the aggregated Grad-CAM maps.  For instance, label (a) corresponds to the aggregated importance map for Soft Cells evaluated through the Position Branch. }
    \label{fig:heatmaps}
\end{figure}

We employ Grad-CAM to visualize how the trained model utilizes input information when making predictions. Specifically, we analyze a model trained using SubSeq on an 80\%/20\% training/testing split, which achieved a testing accuracy of 91\% and a ROC-AUC of 0.96. We first present trajectory-level importance maps for individual test samples and then aggregate the importance scores across the test set to construct spatial importance heatmaps in Figures~\ref{fig:trajectories} and~\ref{fig:heatmaps}, respectively. For each prediction target (soft or hard cells), Grad-CAM is applied separately to the position and velocity branches, yielding four sets of importance maps that reveal how each input modality contributes to the model predictions.

Figure~\ref{fig:trajectories} presents the trajectory-level Grad-CAM importance scores for several representative test trajectories under both the soft- and hard-cell prediction targets. The visualizations show that the model consistently assigns high importance to only selected portions of each trajectory, rather than treating the entire trajectory as equally informative. This indicates that the classifier primarily relies on localized trajectory patterns when making its predictions.

To further investigate how feature importance is distributed across the spatial domain for different input modalities (position and velocity), we construct aggregated spatial importance heatmaps by projecting trajectory points onto the $(x,y)$ plane, partitioning the domain into spatial bins, and averaging the Grad-CAM importance scores within each bin across all test trajectories, as described in Section~\ref{sec:proposed}. Figure~\ref{fig:heatmaps} shows that the trained model does not rely uniformly on the entire microfluidic channel when classifying CTC phenotypes. Instead, the Grad-CAM heatmaps reveal localized regions of high importance, indicating that phenotype-discriminative information is concentrated in specific regions of the channel. This observation supports the hypothesis that the hyperuniform micropost array acts as a physical encoder: only specific cell--micropost and cell--fluid interactions within certain regions produce discriminative trajectory signatures that are critical for classification.

A clear distinction is observed between the position and velocity branches, as illustrated in Figure~\ref{fig:heatmaps}. For both soft and hard cells, the velocity-branch heatmaps exhibit more spatially localized regions of high importance, whereas the position-branch heatmaps display broader regions of influence. This suggests that velocity captures more localized and instantaneous information related to the cell's hydrodynamic response, such as sudden acceleration, deceleration, or changes in direction near a micropost. In contrast, the position branch appears to encode more cumulative and global trajectory information.

Figure~\ref{fig:heatmaps} further suggests that the two phenotypes are distinguished not simply by their overall trajectories, but by how their motion responds locally to the surrounding fluid and microposts. Comparing the Grad-CAM heatmaps for soft and hard cells, we observe that the regions of highest importance differ between the two phenotypes. This indicates that the model learns distinct, phenotype-specific trajectory signatures rather than relying on a single discriminative region within the device. For soft cells, the most influential regions may correspond to locations where cell deformation leads to larger local deviations, smoother navigation around microposts, or delayed recovery of momentum following micropost interactions. In contrast, for hard cells, the important regions may reflect locations where the motion exhibits sharper directional changes or more constrained trajectories. These observations are consistent with the underlying physical intuition: soft cells can deform and therefore tend to follow smoother, more compliant trajectories, whereas hard cells exhibit more rigid, constrained deflection patterns.

\FloatBarrier

\section{Discussion \& Conclusion}
\label{sec:conclu}
In this work, we presented an interpretable and data-efficient deep learning framework for trajectory-based circulating tumor cell (CTC) phenotype classification in a hyperuniform microfluidic device. The proposed framework addresses two important challenges in trajectory-based learning: the limited availability of simulated training data and the lack of transparency in deep neural network predictions. To improve model generalization under data-scarce conditions, we introduced the Subsequence (SubSeq) augmentation strategy, which trains the classifier using randomly sampled contiguous trajectory segments. To improve interpretability, we integrated Grad-CAM to identify the trajectory regions and corresponding physical locations within the microfluidic device that contribute most strongly to phenotype classification.

Experimental results demonstrate that SubSeq consistently improves classification performance across varying training data regimes while providing stable and reliable generalization compared to existing augmentation methods. Interpretability analysis further reveals that accurate predictions rely primarily on localized trajectory segments rather than complete trajectories, providing a physical explanation for the effectiveness of SubSeq. Moreover, the Grad-CAM visualizations show that velocity features contain the strongest predictive information, while positional information provides complementary context that further improves classification performance. Aggregated importance maps additionally indicate that discriminative information is concentrated within specific regions of the microfluidic channel and differs between soft and hard cell phenotypes, suggesting that distinct cell-fluid-structure interactions encode phenotype-specific mechanical signatures.

Beyond improving classification accuracy, this work presents an integrated framework in which the hyperuniform micropost array functions as a physical encoder, transforming intrinsic cellular mechanical properties into trajectory signatures that can be decoded by machine-learning models. By mapping learned importance back to the physical domain, the framework provides mechanistic insight into where and how discriminative information arises, potentially informing more efficient biomedical analysis and future microfluidic device design.

We also acknowledge several limitations of the present study. First, the dataset is derived entirely from numerical simulations and contains a relatively small number of trajectories representing only two mechanical phenotypes. Consequently, the extent to which the trained models generalize to experimentally observed cells and to a broader range of cellular mechanical properties remains uncertain. Second, the current study considers only a single microfluidic device configuration and one convolutional neural network architecture. Future work will therefore focus on validating the proposed framework using experimentally acquired trajectory data, incorporating more diverse cellular characteristics and microfluidic geometries, and evaluating the robustness and effectiveness of SubSeq across different neural network architectures. We will also investigate how model interpretability can be incorporated directly into the optimization of device architectures. More broadly, the proposed combination of data-efficient learning and physically grounded interpretability provides a general framework for trajectory-based analysis in microfluidics and other scientific applications where both predictive performance and mechanistic understanding are important.

\appendix
\section*{Appendix}
\label{appendix:imp}
Training parameters and auxiliary strategies, including learning rate scheduling (e.g. linear warm-up, cosine annealing) and noise injection, are systematically explored for each augmentation method. These techniques are incorporated selectively and only when they are observed to improve performance for a given method during fine-tuning. As a result, each augmentation approach is evaluated under its own best-performing configuration. Hyperparameters and auxillary strategies are tuned independently for each method using the same evaluation protocal, ensuring a fair and consistent basis for comparison.

The SubSeq-augmented model utilizes a learning rate of $3 \times 10^{-4}$, 
an activation probability of $0.99$, and a minimum length ratio of $0.6$ relative to the full sequence length. 

\paragraph{SubSeq against alternative data augmentation methods.}
We consider a fixed $80/20$ train-test split. In this experiment, the \textit{SubSeq} method uses the same parameters as outlined above. Cutout was implemented with a learning rate of $3 \times 10^{-4}$, a masking probability of 0.99, and a maximum mask width of 250. Mixup was implemented with a learning rate of $5 \times 10^{-4}$ and a mixing coefficient of $\alpha = 0.5$.

\section*{Data \& Code Availability}
The CTC trajectory dataset used in this study was obtained from the GitHub repository: \url{https://github.com/imsanjoykb/Microfluidic-Device-Data-CTC_Model}. 
     
The code required to reproduce the results presented in this paper is publicly available at: \url{https://github.com/sjsusu/ctc-classification-model}. All code is released under the MIT License.
\section*{Acknowledgement}
S. L. acknowledges partial support from the Texas Tech University startup funds. Y. W. acknowledges partial support from the National Science Foundation under grant DMS-2247001, the Cancer Prevention and Research Institute of Texas (CPRIT) under grant RP260780, and Simons Foundation Travel Award. The authors thank Dr. Wei Li for insightful discussions regarding circulating tumor cell dynamics and microfluidic device design.
\printcredits

\bibliographystyle{cas-model2-names}

\bibliography{ctc_reference_yifan, ref_serena}

\end{document}